\documentclass[runningheads]{llncs}
\usepackage{graphicx}
\usepackage{comment}
\usepackage{amsmath,amssymb} 
\usepackage{color}
\usepackage{multirow, tabu}
\usepackage{float}
\usepackage{booktabs}
\usepackage{ctable}

\begin{document}
\pagestyle{headings}
\mainmatter
\def\ECCVSubNumber{3451}  

\title{LapNet : Automatic Balanced Loss and Optimal Assignment for Real-Time Dense Object Detection} 

\titlerunning{LapNet}
%
\author{Florian Chabot \and
Quoc-Cuong Pham \and
Mohamed Chaouch}
\authorrunning{F. Chabot, QC. Pham, M. Chaouch}
%
\institute{CEA, LIST, Vision and Learning Lab for Scene Analysis, France \\
\email{florian.chabot@cea.fr}\\
}
\maketitle

\begin{abstract}
Real-time single-stage object detectors based on deep learning \cite{ssd,yolo,yolov2,yolov3} still remain less accurate than more complex ones \cite{retina,fcos}. The trade-off between model performance and computational speed is a major challenge. In this paper, we propose a new way to efficiently learn a single-shot detector which offers a very good compromise between these two objectives (Figure \ref{fig:tradeoff}). To this end, we introduce LapNet, an anchor based detector, trained end-to-end without any sampling strategy. Our approach aims to overcome two important problems encountered in training an anchor based detector: (1) ambiguity in the assignment of anchor to ground truth and (2) class and object size imbalance. To address the first limitation, we propose a soft positive/negative anchor assignment procedure based on a new overlapping function called ”Per-Object Normalized Overlap” (PONO). This soft assignment can be self-corrected by the network itself to avoid ambiguity between close objects. To cope with the second limitation, we propose to learn additional weights, that are not used at inference, to efficiently manage sample imbalance. These two contributions make the detector learning more generic whatever the training dataset. Various experiments show the effectiveness of the proposed approach. 

\end{abstract}

\section{Introduction}
\begin{figure}[ht!]
\center
\includegraphics[width=8cm]{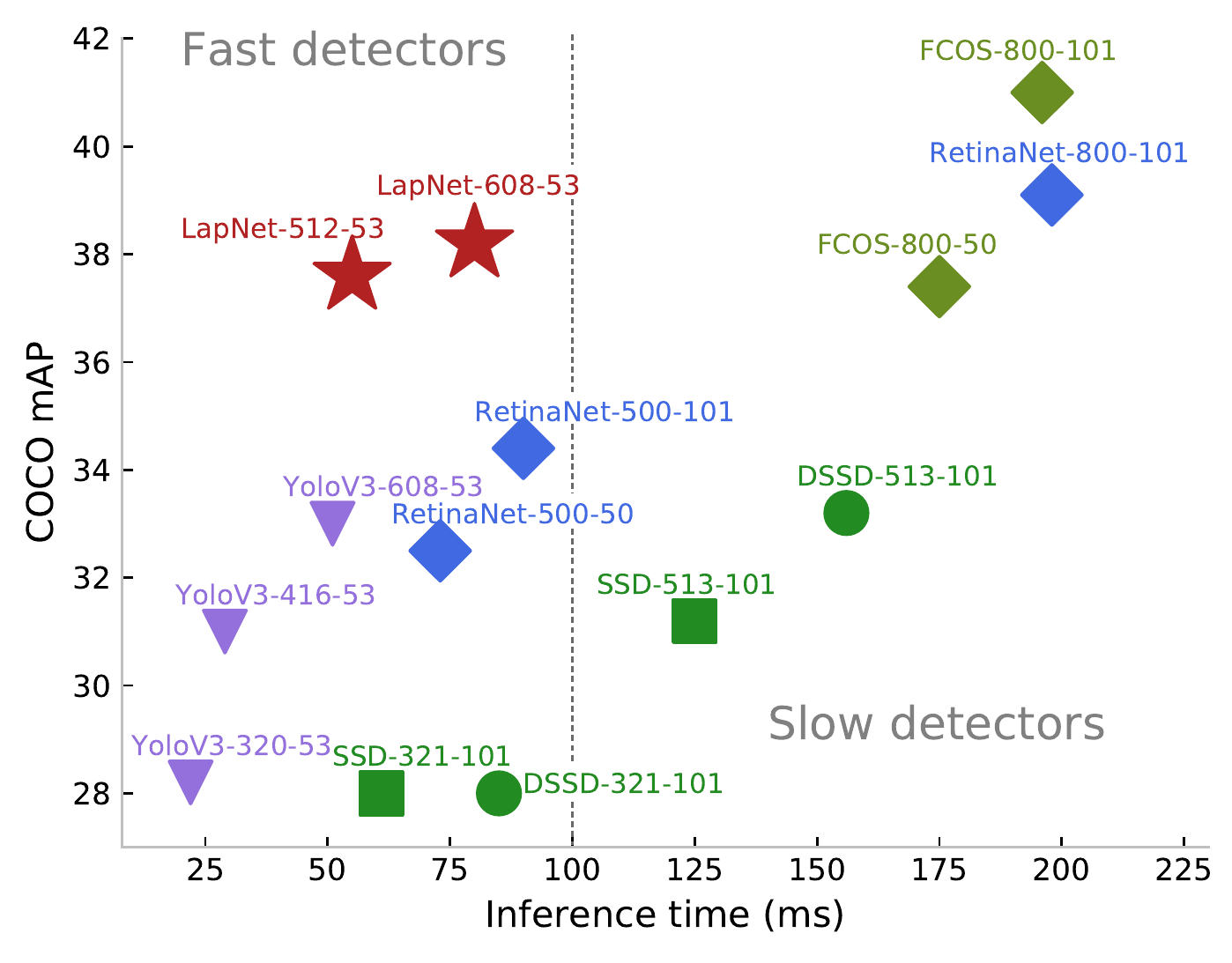}
\vspace{0mm}
\caption{Speed/accuracy trade-off for single-shot detectors on COCO \texttt{test-dev} 2017 using TITAN X GPU. We only plot detectors with a computing time less than 200 ms. Fast detectors correspond to approaches with a processing time per image less than 100 ms. Methods are denoted by "name-resolution-backbone" (50: ResNet-50, 101: ResNet-101, 53: DarkNet-53). LapNet gives significantly better results than YoloV3 \cite{yolov3}, using the same backbone network while keeping the same computing time. With an input image resolution of 608x608, LapNet has similar accuracy than RetinaNet \cite{retina} but a smaller inference time.}
\label{fig:tradeoff}
\end{figure}

Object detection has been widely studied and it still remains a research field of interest for scene understanding applications. It needs to face several challenges: small object detection, heavily occluded objects and real time processing. Modern detectors are based on deep convolutional neural networks (CNN) which have shown their effectiveness in several computer vision tasks (classification, segmentation, detection...). These object detectors can be divided into two main categories: two-stage and single-stage detectors. Two-stage detectors such as \cite{fast,faster} first propose a set of regions of interest in which deep feature maps are cropped and classified. Although these methods achieve good performance, they are time consuming because of their cascaded pipeline. In contrast, single-stage detectors \cite{yolo,yolov2,ssd,retina,fcos} directly predict detection boxes without any region extraction and refinement step. Some of them \cite{yolo,yolov2,yolov3,ssd} are fast and thus suitable for real-time detection, at the cost of a lower accuracy.

Recent object detection algorithms generally use a set of discrete pre-defined boxes called anchors. The network predicts if an anchor contains an object (positive anchor) or not (negative anchor) as well as offsets applied to anchors to fit the objects. The strategy for finding positive and negative anchors during training is based on the absolute overlap criterion using Intersection over Union function (IoU): an anchor is labeled as positive if a ground truth box overlaps it above a fixed threshold. Thresholding the overlap is a way to define what an object is and what it is not. However, this strategy could be discussed. For instance, due to the discrete property of the anchor set, there is some chance that no anchor overlaps a ground truth box above the threshold, especially for small objects. In addition, an anchor highly overlapping two objects is assigned to only one of them. This ambiguous case perturbs the model training.

Another general issue when training object detectors is class imbalance and object size imbalance. Best represented classes (large number of training samples) and large objects are generally best trained. Positive anchors are more easily associated to large objects and dominate the loss function. A standard way to address imbalance is positive/negative sampling strategy \cite{faster,ssd,dssd} based on heuristics. Another way to solve it is to design a specific loss function such as the focal loss \cite{retina} which explicitly manages the balance. The objective of the focal loss is to minimize the influence of easy samples (\textit{i.e} background samples) in the total loss. It balances foreground and background samples in an automatic hard data mining way, but it does not provide a global weighting function which explicitly takes into account class representativeness and object sizes.

In this paper, we introduce LapNet, a new single-stage object detector combining real time processing and efficient detection. LapNet is based on a CNN encoder-decoder architecture which takes as input an image and a set of predefined anchors. It outputs a dense set of bounding boxes with associated per-class probabilities. In this work, we present two main contributions.

First, we propose a new method to choose positive/negative anchors. As stated above, the assignment based on standard absolute overlap (AO) criterion greatly depends of the number of anchors. To overcome this limitation, we introduce Per-Object Normalized Overlap (PONO), presented in section \ref{pono}. PONO is the overlap between an anchor and a given ground truth box, normalized by the maximum overlap between this ground truth and all its assigned anchors. In other words, each ground truth box has at least one anchor which has a PONO value equal to one. It has two advantages compared to the absolute overlap criterion. The first advantage is an increased robustness when applying a threshold to select positive and negative anchors. When a ground truth box has too small AO values, it is discarded. PONO solves this issue by design. The second advantage is that PONO normalizes the overlap score which makes its more insensitive to the anchor discretization. LapNet takes its name from this idea. In addition, a strategy allowing the network to self-correct the positive and negative anchor assignment during training is proposed. The intuition is that a positive anchor could highly overlap almost equally several objects. When such an ambiguity happens, the network will not be able to fit the right object. To solve this issue, we propose to consider an ambiguous anchor as negative using an online Ambiguity Management Strategy (AMS) described in section \ref{training}. 

The second contribution is related to the loss function. Dense optimization for object detection is still difficult because of large imbalance of classes and object sizes. The authors of \cite{homo} proposed the homoscedastic loss for automatic loss balancing in the context of multi-task learning. In our work, we generalize this concept to address general sample imbalance by automatically learn weights which balance classes and object sizes (explained in section \ref{training}). It is a generic alternative to multi-scale detection used in recent litterature \cite{fpn,retina,fcos,yolov3}. These methods use Feature Pyramid Network (FPN \cite{fpn}), which consists in predicting boxes for several feature levels (multi-scale prediction). During training, FPN based approaches balance object sizes in the network architecture itself: large objects are optimized in the lower pyramid level and small objects in the higher pyramid level. It forces to define which anchors correspond to each level. Differently, LapNet's generic weight learning avoids the use of multi-scale detection.  

The two contributions of this work lead LapNet to be comparable to best performing object detectors with a significant reduction of the processing time. This is detailed in section \ref{exp}.

\section{Related work}

Given an input image, an object detector returns a list of bounding boxes and associated confidence score for each class. An object detector classifies image regions to predict those containing an object. A standard way to analyse all regions in the image is the sliding window scheme which has been widely used by reference work \cite{viola,dpm,hog}. With the advent of Deep Learning in computer vision community, sliding window and hand-crafted features for object detection have been surpassed by CNN based methods. \\ \\
\textbf{Anchor based detectors.} Most CNN based methods \cite{faster,ssd,yolov2,retina} use anchors to detect objects. Anchors are pre-defined boxes that are supposed to be representative of all objects in terms of scale and ratio. Each 2D position in the CNN output map corresponds to one anchor at this position. The objective is to predict if an anchor contains an object or not, as well as the offsets to apply to the anchor to fit the object. During training, positive and negative labels are assigned to anchors by using the absolute overlap criterion. Classification loss (cross entropy, focal loss) and regression loss (smoothL1) functions are used to train the model. 
One limitation of these methods is related to the positive/negative anchor assignment. Because of the absolute overlap criterion, some ground truth boxes can be left out. In this paper, we propose to solve this issue using a soft overlap criterion and an Ambiguity Management Strategy allowing to be less sensitivity to the number of pre-defined anchors. \\ \\
\textbf{Two-stage vs single-stage detectors.} The first two-stage based detector was proposed by the authors of \cite{rcnn}. They introduced the object proposal concept: a multi-scale segmentation algorithm \cite{ss} is executed to propose several candidate boxes. These regions are then extracted in the image and forwarded into a CNN that predicts class scores and regression offsets. This work was extended in \cite{fast} where deep feature maps are extracted in regions of interest (using ROI Pooling) to be more efficient and faster. In 2015, Faster R-CNN \cite{faster} was introduced: it combines both anchor based object proposal generation and region classification in the same deep network. Since, Faster R-CNN has been well studied and improved \cite{fpn,fcn,dm} making it the reference method for two-stage detection. However, cascaded methods such as Faster R-CNN are time consuming in inference phase. For this reason, single-stage methods were first introduced with the SSD algorithm \cite{ssd}. The network directly outputs a dense set of boxes and associated per class scores. Since, many single-shot approaches have been developed  \cite{yolo,yolov2,dssd,retina,fcos}. The main advantage of these methods is shorter processing time: with a reasonable input resolution and a light backbone network, some of these detectors give access to real time processing \cite{ssd,yolo,yolov2,yolov3}. However, there still exists an important performance gap between fast single-shot detector and heavier ones such as RetinaNet \cite{retina} or the recent FCOS \cite{fcos}. With LapNet, we provide high mean Average Precision (mAP) while keeping real time processing.   \\ \\
\textbf{Sample imbalance.} Class imbalance is a well known issue which is not specific to object detection. For instance, Dice loss \cite{dice} is often used in medical image segmentation to address large class imbalance. In object detection, object scales and class representativeness are two important things to take into account. To not focus on large objects, some approaches use sampling strategy \cite{faster,ssd,dssd} or bootstrapping \cite{ohem}. In \cite{retina,fcos}, the focal loss automatically balances foreground and background samples by giving more weight to hard samples. In our work, a generic method for both classes and object scales imbalance is proposed. To our knowledge, no method modeling sample balance with trainable parameters exists. Our approach generalizes the homoscedastic loss proposed by \cite{homo} in the context on multi-task learning.

\section{LapNet}

\subsection{Overview} 

\begin{figure*}[t]
\center
\includegraphics[width=12cm]{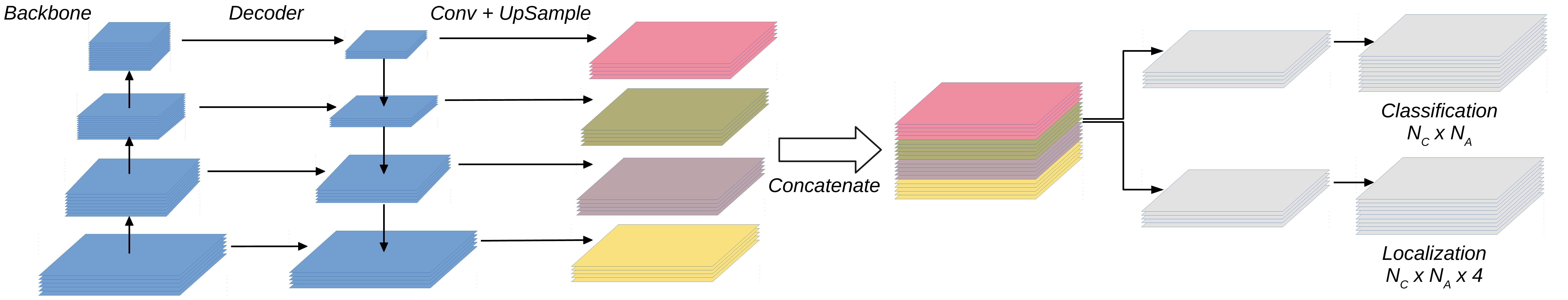}
\vspace{2mm}
\caption{LapNet encoder-decoder architecture overview. The input image is passed through a backbone network. A decoder with skip connections is then used to compute feature maps at several resolutions. Each of these feature maps is then processed by four 3x3 convolutional layers and resized to the size of the highest resolution feature map. After multi-scale feature concatenation, we use two convolutional heads to provide the two outputs. Each head is composed by four 3x3 convolutional layers and one final convolutional layer returning the output.  }
\label{fig:overview}
\end{figure*}

Given an input image, LapNet predicts a dense set of bounding boxes for each class and the associated confidence scores. These boxes are then filtered using non-maximum suppression to provide final detections. As in \cite{faster,yolov2,fpn,ssd}, LapNet uses anchors corresponding to pre-defined boxes. More specifically, the network returns for each anchor (1) the probability that it contains an object of a given class and (2) the offsets to apply to fit the object. \\ \\
\textbf{Per-class anchors.} Anchors are computed on the training dataset with the K-means clustering method proposed in \cite{yolov2}. Contrary to related work, anchors are computed for each class independently. In this way, anchors are more specific and better fit the distribution of size and aspect ratio of objects for a given class. This is especially true in several contexts as in traffic scene analysis where the classes "car" and "person" do not share the same box distribution.
We define $N_C$ as the number of classes and $N_A$ the number of anchors for each class. To simplify, $N_A$ is the same for each class. With this formulation, LapNet returns $N_C \times N_A$ score maps and $N_C\times N_A \times 4$ box offsets (corresponding to the four offsets values $dx, dy, dw, dh$ to apply to an anchor to fit its associated object as in \cite{fast,faster,retina}). \\  \\
\textbf{Architecture.} LapNet is a fully convolutional network based on an encoder-decoder architecture with skip connections. This architecture allows to keep information of objects at multiple resolutions and contextual information. Unlike FPN based approaches \cite{fpn,retina,yolov3}, detection is not performed on each pyramid level. Instead, each pyramid feature map is passed through four convolutional layers and resized to the size of the highest resolution level. These feature maps are then concatenated and forwarded into a classification head and an offset regression head to predict the network outputs. In addition to its simplicity, the advantage of this architecture resides in its genericity compared to multi-scale detection based methods. Indeed, it does not require an explicit assignment between anchors and pyramid levels during training. Figure \ref{fig:overview} illustrates the LapNet architecture.

\subsection{Dense association between anchors and ground truth boxes}
\label{pono}
LapNet requires to define labels on the entire anchor set. In previous work \cite{faster,retina,ssd}, anchors with positive or negative label are defined according to a threshold applied to the absolute overlap (AO) criterion: if the overlap value of an anchor with a ground truth box is above the threshold, the anchor is considered as positive. The main drawback of absolute overlap measure is that some ground truth boxes may be missed by the label assignment because of the discrete property of anchor set. This usually happens in the case of small and strongly occluded objects.

\begin{figure}[t]
\center
\includegraphics[width=6cm]{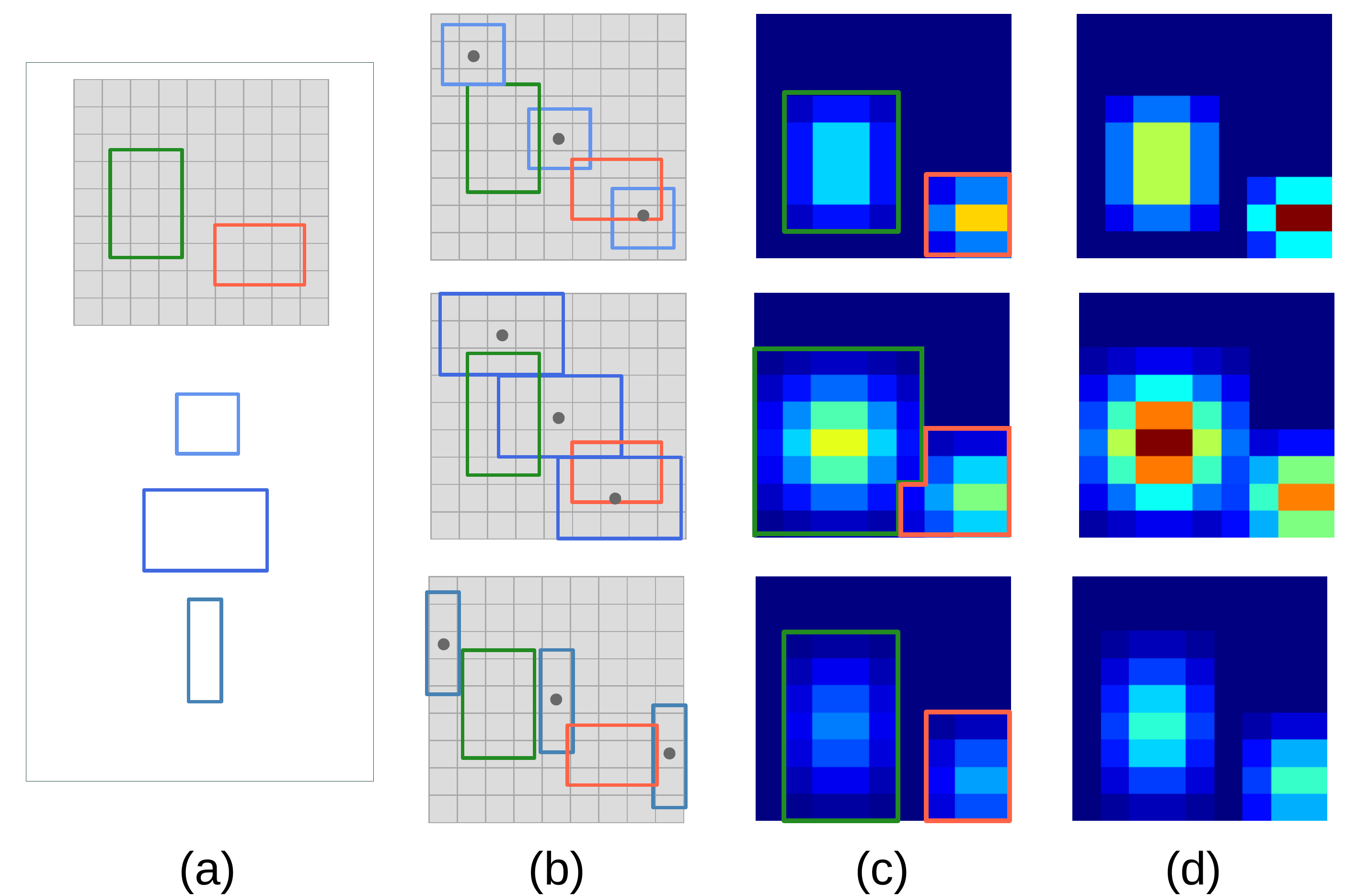}
\vspace{0mm}
\caption{Per-Object Normalized Overlap (PONO) computation. (a) An input image with two ground truth objects (top) and three pre-defined anchors (bottom). (b) Dense anchor grid representation. (c) Anchor to ground truth assignment using absolute overlap, cluster colors (green and orange) correspond to the ground truth object assigned to each anchor of the grid. (d) Per-Object Normalization Overlap. The two ground truth boxes have at least one anchor with a PONO value equal to one (in red).}
\label{fig:anchors}
\end{figure}

To solve this issue, we introduce a new overlap measure called Per-Object Normalized Overlap function (PONO). It is a relative measure that increases the value of AO, especially since the AO value is low. Additionally, it always gives an overlap value equal to one for at least one anchor associated to a given object. Consequently, no object in the ground truth is missed. \\ \\
\textbf{Anchor to ground truth assignment.} A grid of anchors is computed using pre-defined anchors (Figure \ref{fig:anchors}b). It consists in a grid of size $H_f \times W_f \times N_C \times N_A \times 4$  where $H_f$ and $W_f$ are the height and width of the output feature map. $N_A$ and $N_C$ are respectively the number of anchors per class and the number of classes. Each element of the anchor grid is assigned to a ground truth box using the AO criterion. This assignment forms clusters where each anchor of the cluster corresponds to the same object in the ground truth (Figure \ref{fig:anchors}c). \\ \\
\textbf{Per-Object Normalized Overlap.} Assignment clusters are then used to compute the PONO map $O$. For each cluster, the highest overlap is selected. Each element of the cluster is divided by its corresponding maximum. Let $A_{c,a,i,j}$ be the anchor of class $c$, of index $a$ at position $(i,j)$ in the feature map.
For a given ground truth box $B_n$, we denote  $\mathcal{C}_{B_n}$ the set of anchors associated to $B_n$. The PONO function is then defined by:

\begin{equation}\label{eq:pono}
O(A_{c,a,i,j}, B_n) = \frac{IoU(A_{c,a,i,j}, B_n) }{\max\limits_{A_{c,a^\prime,i^\prime,j^\prime} \in \mathcal{C}_{B_n}} IoU(A_{c,a^\prime,i^\prime,j^\prime},  B_n)} \text{\quad with \quad} A_{c,a,i,j} \in \mathcal{C}_{B_n}
\end{equation}


In the following, we consider the resulting PONO map $O = \{O_{c,a,i,j} = O(A_{c,a,i,j}, B_n)\}$.
By definition, $O$ has at least one element of value 1 for each object (Figure \ref{fig:anchors}d). It will be used as input in the training process described in the next part. 

\begin{figure*}[t]
\center
\includegraphics[width=10cm]{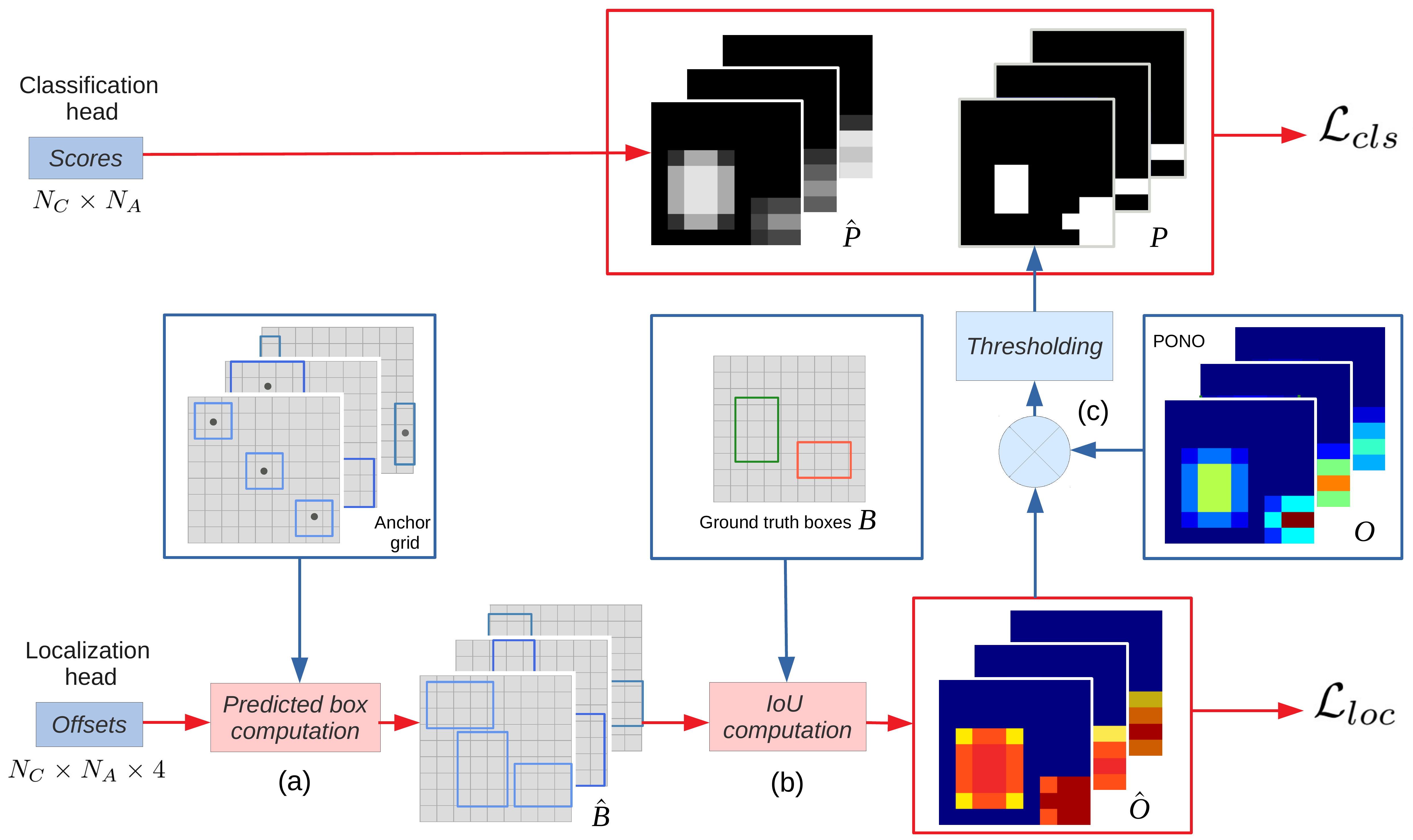}
\vspace{1mm}
\caption{LapNet training process. Given an image, the network outputs scores and offsets. (a) Offsets are applied to the anchor grid to get predicted boxes. (b) IoU map $\hat{O}$ is then computed using predicted boxes and associated ground truth boxes. $\hat{O}$ is used to compute the localization loss $\mathcal{L}_{loc}$. (c) The thresholded product between PONO map $O$ and predicted IoU map $\hat{O}$ determines classification labels to optimize the classification loss $\mathcal{L}_{cls}$. Back-propagation is performed on the inverse path of flows colored in red.}
\label{fig:training}
\end{figure*}

\subsection{LapNet training process}
\label{training}

To train LapNet, two loss functions are used: a localization loss $\mathcal{L}_{loc}$, and a classification loss $\mathcal{L}_{cls}$. They use the pre-computed PONO map $O$ described in the previous section. The localization loss is used to regress offsets applied to an anchor to fit its associated ground truth box. The classification loss is used to predict if an anchor contains an object of a specific class. In this work, we propose a weighted variant of these losses to avoid class and object size imbalance. Manually finding weights or standard hyper-parameters search to balance samples or losses for better training is time consuming. To deal with this issue, we propose an efficient optimization strategy to find these weights automatically in the training loop, without extra cost in computing time. The overall training process is illustrated in Figure \ref{fig:training}. 

\subsubsection{Pixel-wise localization loss.}

In LapNet, the localization loss uses the Intersection over Union function to train the box offset regression task. Contrary to previous anchor based approaches \cite{faster,yolov2,yolo,retina} which directly learn to regress box offsets using SmoothL1 loss, we propose to train the model to predict box offsets in a latent way, by optimizing the IoU function. It is a bounded function that is more stable when predicted offsets have high values. This is particularly true when using PONO because it recovers positive anchors associated to objects that are difficult to detect.

First, by applying predicted offsets to the anchors, predicted boxes are generated (Figure \ref{fig:training}a). Then, the IoU $\hat{O}_{c,a,i,j}$ between a predicted box and its assigned ground truth box is computed. $\hat{O}$ is the resulting IoU map (Figure \ref{fig:training}b). Finally, using the PONO map $O$, the localization loss function is defined as follows:

\begin{equation}
 \mathcal{L}_{loc}(c, a, i, j)=
    \begin{cases}
      \lVert 1 - \hat{O}_{c,a,i,j}) \rVert^2, & \text{if}\ O_{c,a,i,j} > 0.5\\
      0, & \text{otherwise}
    \end{cases}
\end{equation}

This loss function allows to learn latent offsets so that the overlap beetween a predicted box and its associated ground truth box is equal to one. The IoU loss  has already been used for pixel-to-box regression \cite{fcos,unit} but not for anchor based detector.

\subsubsection{Pixel-wise classification loss with label ambiguity management.}
\label{Pixel-wise classification loss}

\begin{figure*}[t]
\center
\includegraphics[width=12cm]{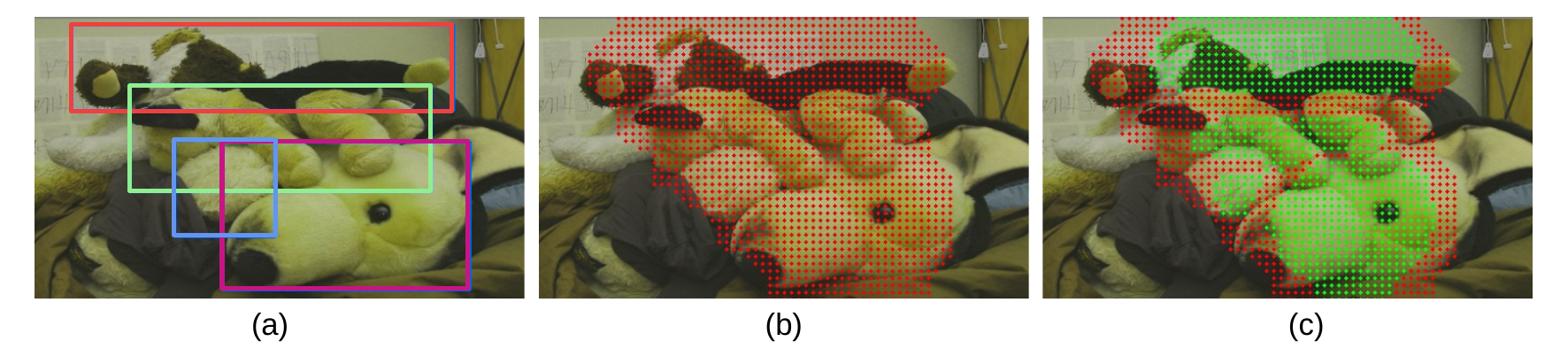}
\vspace{2mm}
\caption{Illustration of ambiguity management (best viewed in color). (a) An image with four overlapping ground truth bounding boxes. (b) Selected positive anchors by thresholding PONO values. For readability, anchors are represented by red points. (c) Selected positive anchors (green points) by thresholding PONO map $O$ weighted by geometric overlap $\hat{O}$. We observe that green points form four clusters corresponding to the four objects. Remaining red points correspond to ambiguous anchors (on object borders) labeled as negative anchors.}
\label{fig:ambiguity}
\end{figure*}

A standard way to define labels for classification loss would be thresholding the PONO map as it is done in the localization loss. Although PONO copes with the problem of unassigned ground truth objects, there still exists the issue of ambiguity between objects that are spatially close in the image. Indeed, an anchor that overlaps almost equally several objects is likely to fit any of these objects. Such an anchor is ambiguous and should not be considered as a positive anchor during training because it is generally unable to fit its assigned object correctly. Figure \ref{fig:ambiguity} illustrates the ambiguity situation.

To handle such cases, we propose a strategy for determining anchor labels by using the product of the PONO map $O$ by the predicted geometric overlap $\hat{O}$ (Figure \ref{fig:training}c). Using this strategy, positive and negative labels are defined as follows:
\begin{equation}
 P_{c, a, i, j}=
    \begin{cases}
      1, & \text{if}\ O_{c,a,i,j} \times  \hat{O}_{c,a,i,j} > 0.5\\
      0, & \text{otherwise}
    \end{cases}
\end{equation}
In case of ambiguity, if the network cannot predict offsets correctly for a given anchor, the value of $\hat{O}_{c,a,i,j}$ will be low and therefore the product will return a low score potentially below the threshold. Such an anchor will then be considered as a negative one. Finally, the pixel-wise classification loss is defined by:
\begin{equation}
\mathcal{L}_{cls}(c, a, i, j) = CE(P_{c, a, i, j}, \hat{P}_{c, a, i, j})
\end{equation}
where $CE$ is the standard binary cross-entropy and $\hat{P}_{c, a, i, j}$ is the probability that the anchor $A_{c, a, i, j}$ contains an object of the class $c$. The sigmoid activation function is applied on logits to get $\hat{P}$.

\subsubsection{Balanced losses.}

One important challenge in training multi-class models for object detection or semantic segmentation is class and object scale imbalance. Underrepresented classes are much less well learned than well represented classes. Additionally, large objects dominate the loss function leading to poor performance on small objects. To solve it, several methods use sampling strategies \cite{faster,fpn}, multi-scale detection head \cite{fpn,retina,fcos,yolov3}, hand-crafted weights \cite{pspnet,segnet} or specific loss \cite{retina}. 

In our work, we propose to automatically learn balance weights. Inspired by \cite{homo}, that addresses loss balancing for multi-task training, we learn two kinds of weights: loss weights and anchor weights. Loss weights are used to balance the different loss functions in the total loss. Our main contribution here, is the anchor weights intended for balancing classes and object sizes. More formally, we propose the following weighted losses:
\vspace{0mm}
\begin{equation} \label{eq:classification}
\mathcal{L}_{loc} = \lambda_{loc} \frac{1}{N^+} \sum_{c} \sum_{a} \lambda_{loc}^{c,a}  \sum_{i,j} \mathcal{L}_{loc}(c, a, i, j)
\end{equation}
\vspace{-3mm}
\begin{equation} \label{eq:localization}
\mathcal{L}_{cls} = \lambda_{cls} \frac{1}{N} \sum_{c} \sum_{a} \lambda_{cls}^{c,a}  \sum_{i,j} \mathcal{L}_{cls}(c, a, i, j)
\end{equation}
where $\lambda_{cls}$, $\lambda_{loc}$ are the loss weights and $\lambda_{cls}^{c,a}$, $\lambda_{loc}^{c,a}$ are the anchor weights, defined for a given anchor $a$ and a class $c$. $N^+$ is the number of positive anchors for the localization task and $N$ is the total number of pixels. In this formulation, all $\lambda$ weights are trainable variables. A regularization loss $\mathcal{L}_{reg}$ is added to avoid $\lambda=0$:
\begin{equation}
\begin{split}
\mathcal{L}_{reg} & = log(\frac{1}{\lambda_{cls}}) + log(\frac{1}{\lambda_{loc}}) \\ 
& +\frac{1}{N_C N_A}\sum_{c} \sum_{a} log(\frac{1}{\lambda_{cls}^{c,a}}) + log(\frac{1}{\lambda_{loc}^{c,a}})  
\end{split}
\end{equation}

In practice, instead of directly optimizing $\lambda$, we use  $s = log(\frac{1}{\lambda}$) as trainable variables for better numerical stability and to avoid weights to be negative. In addition, if an anchor grid does not have a least one positive label, corresponding anchor weights $\lambda_{cls}^{c,a}$ and $\lambda_{loc}^{c,a}$ are not updated. This is important to avoid huge values of weights which could make the network diverge. In fact, the objective of anchor weights is to balance the loss of each anchor grid. If an anchor grid has only negative samples, it is an easy case for the network and thus its loss will be very small. In this case, the anchor weight increases too much compared to the other weights of anchor grids containing positive samples.
Finally, the total loss for training LapNet is the sum of the localization, classification and regularization losses. 

\section{Experiments}
\label{exp}

In this section, the effectiveness of LapNet is demonstrated through various experiments. In particular, we provide results on two challenging datasets (PASCAL VOC \cite{voc}, MSCOCO \cite{coco}). Ablation studies are conducted to show the relevance of each contribution.  \\ \\
\textbf{Training details.} In all experiments, SGD with momentum of 0.9, initial learning rate of 0.005 and polynomial decay policy with power of 0.9 are used. For data augmentation, random scale and horizontal random flipping are applied. We arbitrarily fix the number of anchors per object to $N_A=10$. The trainable variables $s=log(\lambda)$ are all initialized to 1 (we observed that the initial value does not influence final performances). Two pre-trained backbone networks, DarkNet-53 \cite{yolov3} and Inception-Resnet-V2 \cite{incres} are used, both pre-trained on ImageNet \cite{imagenet}. The training is performed on 4 GPUs with a global batch size of 28 for DarkNet-53 and 16 for Inception-Resnet-V2. For DarkNet-53, the decoder structure proposed in YoloV3 \cite{yolov3} is used and for Inception-Resnet-v2, the decoder structure proposed by \cite{retina} is plugged into the backbone. In both cases, the spatial output size of the network is eight times smaller than the input image (feat stride = 8).  

\subsection{PASCAL VOC}

This dataset contains 20 classes of objects. LapNet is trained using the union of 2007 and 2012 \texttt{trainval}, and tested on 2007 \texttt{test}. 120K iterations are runned to get the final result shown in Table \ref{results_voc07_test}. We can see that LapNet improves detection results compared to previous work for both backbone networks. In particular, LapNet, trained end-to-end, offers better performance than the best previous method DSSD \cite{dssd} which uses multiple training stages to get the final detector. In the following, an ablation study based on PASCAL VOC is presented.

\subsubsection{PONO and ambiguity management strategy.}
In this part, we evaluate the relevance of Per-Object Normalized Overlap and ambiguity management strategy. The analysis is presented in Table \ref{abation_result}a.
All models are trained using loss weights, anchor weights and the DarkNet-53 backbone \cite{yolov3}. In the first column of Table \ref{abation_result}a, we compare the performance of LapNet models trained with PONO and the standard absolute overlap (AO) label assignement criteria. We observe better detection results are obtained with PONO than with absolute overlap (81.1 vs 80.3), which demonstrates that PONO is a better criterion for positive/negative label assignment. The last column of Table \ref{abation_result}a provides results when ambiguity managment strategy, described in \ref{Pixel-wise classification loss}, is added to choose positive and negative anchors. Combining ambiguity management with AO decreases the mAP (78.4 vs 80.3). This could be explained by the fact that AO already misses some ground truth objects. Ambiguity management further deteriorates the result because it makes the overlap value lower. Combining PONO and ambiguity management gives the best result (81.7 vs 81.1).

\begin{table}[t]
\centering
\scalebox{0.67}{%
\scriptsize
\begin{tabular}{l|l|c|cccccccccccccccccccc}
method & backbone & mAP & areo & bike & bird & boat & bottle & bus & car & cat & chair & cow & table & dog & horse  & mbike & person & plant & sheep & sofa & train & tv \\ \hline
Faster\cite{faster} & VGG-16 & 73.2 & 76.5 & 79.0 & 70.9 & 65.5 & 52.1 & 83.1 & 84.7 & 86.4 & 52.0 & 81.9 & 65.7 & 84.8 & 84.6 & 77.5 & 76.7 & 38.8 & 73.6 & 73.9 & 83.0 & 72.6 \\
Faster\cite{faster} & ResNet-101 & 76.4 & 79.8 & 80.7 & 76.2 & 68.3 & 55.9 & 85.1 & 85.3 & 89.8 & 56.7 & 87.8 & 69.4 & 88.3 & 88.9 & 80.9 & 78.4 & 41.7 & 78.6 & 79.8 & 85.3 & 72.0 \\
MR-CNN\cite{mrcnn} & VGG-16 & 78.2 & 80.3 & 84.1 & 78.5 & 70.8 & 68.5 & 88.0 & 85.9 & 87.8 & 60.3 & 85.2 & 73.7 & 87.2 & 86.5 & 85.0 & 76.4 & 48.5 & 76.3 & 75.5 & 85.0 & 81.0 \\
R-FCN\cite{fcn} & ResNet-101 & 80.5 & 79.9 & 87.2 & 81.5 & 72.0 & \textbf{69.8} & 86.8 & 88.5 & 89.8 & \textbf{67.0} & 88.1 & 74.5 & 89.8 & 90.6 & 79.9 & 81.2 & 53.7 & 81.8 & 81.5 & 85.9 & 79.9 \\ \hline
YOLOv2\cite{yolov2} & DarkNet-19 & 78.6 & - & - & - & - & - & - & - & - & - & - & - & - & - & - & - & - & - & - & - & - \\
SSD300\cite{ssd} & VGG-16 & 77.5 & 79.5 & 83.9 & 76.0 & 69.6 & 50.5 & 87.0 & 85.7 & 88.1 & 60.3 & 81.5 & 77.0 & 86.1 & 87.5 & 83.97 & 79.4 & 52.3 & 77.9 & 79.5 & 87.6 & 76.8 \\
SSD512\cite{ssd} & VGG-16 & 79.5 & 84.8 & 85.1 & 81.5 & 73.0 & 57.8 & 87.8 & 88.3 & 87.4 & 63.5 & 85.4 & 73.2 & 86.2 & 86.7 & 83.9 & 82.5 & 55.6 & 81.7 & 79.0 & 86.6 & 80.0 \\
SSD513\cite{dssd} & ResNet-101 & 80.6 & 84.3 & 87.6 & 82.6 & 71.6 & 59.0 & 88.2 & 88.1 & 89.3 & 64.4 & 85.6 & 76.2 & 88.5 & 88.9 & 87.5 & 83.0 & 53.6 & 83.9 & 82.2 & 87.2 & 81.3 \\
DSSD513\cite{dssd} & ResNet-101 & 81.5 & 86.6 & 86.2 & 82.6 & 74.9 & 62.5 & 89.0 & 88.7 & 88.8 & 65.2 & 87.0 & 78.7 & 88.2 & 89.0 & 87.5 & 83.7 & 51.1 & 86.3 & 81.6 & 85.7 & 83.7 \\ \hline
LapNet512 & DarkNet-53 & 81.7 & 88.1 & 88.7 & 82.0 & 75.0 & 65.8 & 88.1 & \textbf{91.7} & 90.0 & 65.1 & 85.9 & 77.4 & 88.5 & 90.8 & 86.1 & \textbf{86.2} & 51.5 & 82.7 & 81.8 & 89.2 & 79.4  \\
LapNet512 & IncResV2 & \textbf{83.2} & \textbf{89.8} & \textbf{89.8} & \textbf{83.8} & \textbf{76.1} & 65.2 & \textbf{89.8} & 90.7 & 92.0 & 64.6 & \textbf{89.8} & \textbf{79.0} & \textbf{91.8} & \textbf{91.8} & \textbf{89.5} & 84.9 & \textbf{53.5} & \textbf{86.3} & \textbf{82.4} & \textbf{89.6} & \textbf{84.7}  \\
\end{tabular}}
\vspace*{2mm}
\caption{Results on the VOC 2007 \texttt{test} set. Two-stage detectors are on the top of the table and single-stage at the end. Compared to both detector families, LapNet gives the best results.}
\vspace*{-5mm}
\label{results_voc07_test}
\end{table}

\begin{table}[]
    \begin{minipage}{.42\textwidth}
      \centering
      \begin{tabular}{p{20mm}|p{10mm}|p{10mm}}
			AMS  & no & yes   \\ \specialrule{.1em}{.05em}{.05em} 
			AO & 80.3 & 78.4 \\
			PONO & 81.1 & \textbf{81.7} \\ 
	\end{tabular}
    \end{minipage}
    \begin{minipage}{.4\textwidth}
    \centering  
      \begin{tabular}{c|c|c|c|c|c}
$\lambda_{loc}$ & $\lambda_{cls}$ & $\lambda_{loc}^{c,a}$ & $\lambda_{cls}^{c,a}$ & Loss & mAP \\ \specialrule{.1em}{.05em}{.05em} 
 1 & 1 & 1 & 1 & CE & 16.4 \\
 1 & $R$ & 1 & 1 & CE & 79.0 \\
 1 & $R$ & 1 & 1 & FL & 71.4 \\
learned & learned & 1 & 1 & CE & 81.3 \\
learned & learned & learned & learned & CE & \textbf{81.7} \\
\end{tabular} 

    \end{minipage}
    
    \vspace*{1mm}
        \begin{tabular}{c}
\qquad \qquad \qquad \quad (a) \qquad \qquad \qquad \qquad \qquad \qquad \qquad \qquad (b)
	\end{tabular}
	\vspace*{2mm}
    \caption{(a) Influence of PONO and ambiguity management strategy (AMS). AO is the standard absolute overlap. The bottom right result corresponds to the complete LapNet model which outperforms other kind of positive/negative assignment. (b) Influence of automatic weights learning in the loss equation (\ref{eq:classification}) and (\ref{eq:localization}). We provide results for different weighing strategies (hand-crafted fixed weights or learned weights). $R = \frac{N}{N^+}$ represents the normalization used for classication loss in RetinaNet \cite{retina}, see text for detail.  CE is the cross entropy and FL the focal loss. The second line is considered as the baseline. The last line corresponds to the proposed full automatic weighting strategy.}
\label{abation_result}
  \end{table}

\subsubsection{Automatic balancing.} 
\begin{figure}[t]
\center
\includegraphics[width=7cm]{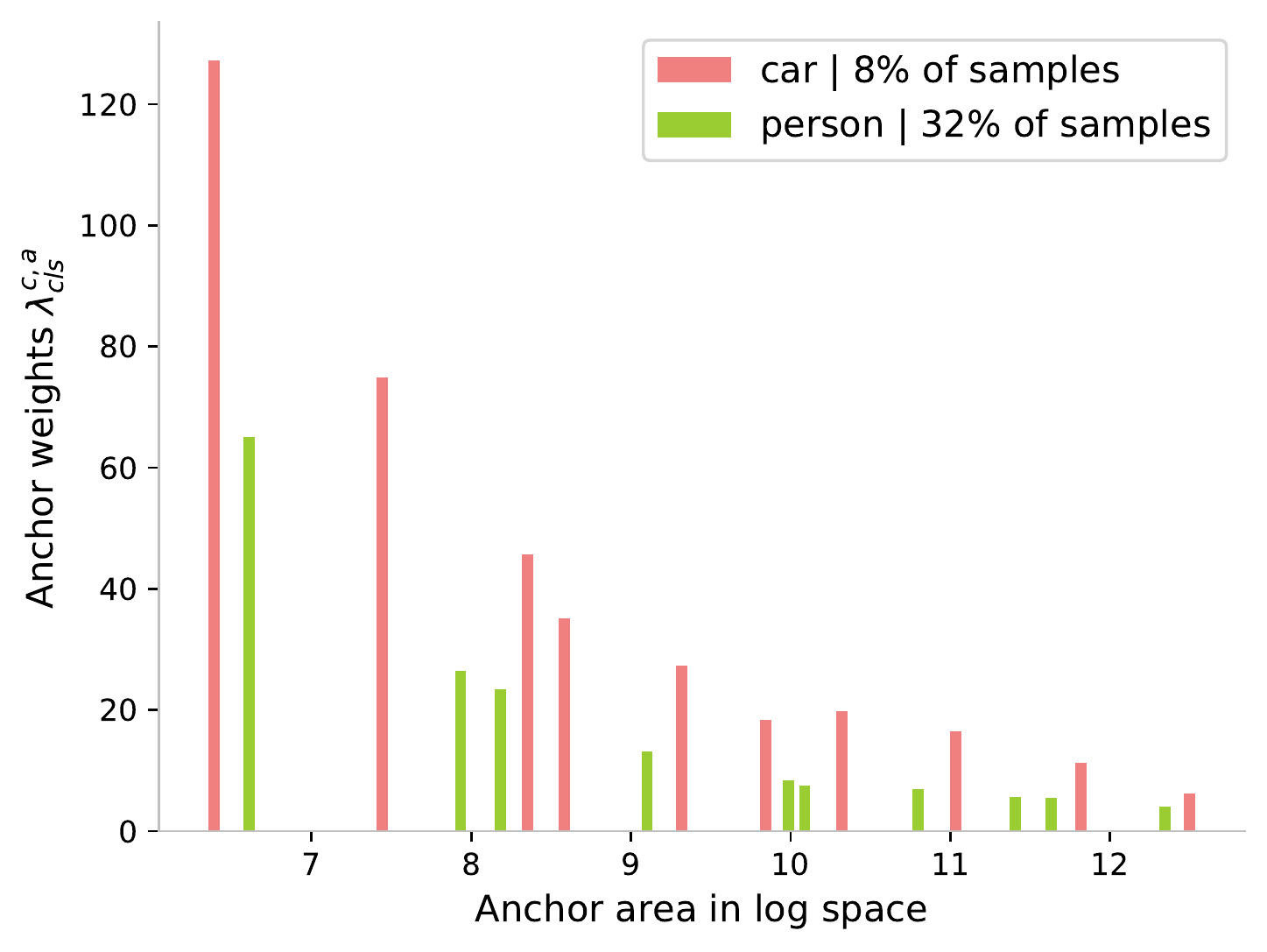}
\vspace{-1mm}
\caption{Analysis of weights automatically learned on VOC2007-2012 \texttt{trainval}. Anchor weights of two classes (car and person) are represented ($N_A=10$). The proportion of samples over the entire training dataset is reported in the top-right corner. This plot shows that weight values decrease with the anchor size and decrease with the number of samples of the class. }
\label{fig:weights}
\end{figure}

We also study the influence of automatic weight learning for loss and anchor balance. Table \ref{abation_result}b summarizes obtained results. All models are trained using the PONO map and the ambiguity management strategy. The first row is a model where weights are constant and equal to one. We observe that the model completely failed as it was expected. The main reason comes from the normalizer $N$ in the equation of the classification loss (\ref{eq:classification}) which is an important parameter. As a matter of fact, $N$ is simply the number of pixels (anchors) in the outputs. In other words, we just take the average over all pixels without any focus on positive anchors unlike RetinaNet \cite{retina}. The resulting classification loss value is therefore very low because of the large number of easy negative samples leading the network to not be trained correctly. In the second row, weights are also constant but we use the normalization method used in \cite{retina} where the classification loss is divided by the number of positive anchors $N^+$. With this formulation and equation (\ref{eq:classification}), we give the constant value $\lambda_{cls} = R = \frac{N}{N^+}$. This experiment is the baseline for our study. In the third line, we repeat the same experiment using focal loss \cite{retina} (with default parameters $\alpha = 0.25$ and $\gamma=2$) instead of cross entropy. Suprisingly, the focal loss decreases results significantly. It may be necessary to tune focal loss hyper-parameters to achieve better performance but it is a time consuming process which is specific to each training dataset. The fourth line of Table \ref{abation_result}b shows that when using only automatic loss weights the model outperforms the baseline model of the second line. That proves that during training, even if classification loss values are low because of dividing by $N$, automatic loss weights naturally counterbalance that phenomenon. Finally the best result is obtained (Table \ref{abation_result}b, last line) when using our proposed full automatic loss and anchor balancing. 
Figure \ref{fig:weights} plots the values of some of the learned anchor weights. This plot shows that the larger is the area of an anchor, the smaller is its learned weight. Besides, the more the class is underreprensented, the larger are the weights. This is the expected behavior which would be hardly designed manually. Thanks to its automatic balancing method, LapNet provides state-of-the-art results on PASCAL VOC.  


\subsection{MSCOCO}
We train LapNet on the 80 object categories using the 2017 \texttt{trainval} split (115K images) and test on 2017 \texttt{test-dev} (20K images) by uploading results on the evaluation server. Results presented in Table \ref{results_coco} are obtained using 500K training iterations. The performance of LapNet is compared to several other state-of-the-art object detectors among which fast single-shot detectors (\textit{i.e} running in less than 100 ms) \cite{yolov3,yolov2,ssd}, slower single-shot detectors \cite{retina,fcos,dssd,cornernet} and two-stage detectors \cite{tdm,fpn,speedAccuracy}. 

The main result is that LapNet widely outperforms the best of fast detectors, YoloV3 \cite{yolov3}, using the same backbone network and the same decoder (37.6 of mAP vs 33.0). This important improvement comes from our proposed learning strategy showing that with almost the same computing time (Figure \ref{fig:tradeoff}), LapNet gives better detection results. 

Compared to stronger but more time consuming methods such as RetinaNet \cite{retina} and FCOS \cite{fcos} taking input images of 800 pixels for the smaller dimension (denoted RetinaNet-800 and FCOS-800), and using ResNet-101-FPN backbone, LapNet is less accurate (-0.9 to -2.8 in mAP) but is 2.5x faster (Figure \ref{fig:tradeoff}). In previous work, the best results of RetinaNet and FCOS were obtained with the heavier ResNeXt-101 \cite{resnext} backbone,  outperforming LapNet by 2.6 and 3.9 points in mAP respectively. They are mentioned in Table \ref{results_coco} for information. However, we do not consider them in the speed/accuracy trade-off analysis because they are not suitable for real-time applications. For the same reason, we did not implement LapNet with ResNeXt-101 backbone. ResNet-101 was not implemented either because it was shown in \cite{yolov3} that DarkNet-53 has equivalent performance. 

Input image resolution is also an important parameter. LapNet is implemented with the same input image resolution as YoloV3 (512x512 or 608x608) to be fast. Compared to the version of RetinaNet using similar input image resolution (500 pixels for the smallest dimension) denoted RetinaNet-500, LapNet clearly achieves better performance (+3.2 in mAP at least, see Figure \ref{fig:tradeoff} and Table \ref{results_coco}). 


\begin{table*}[t]
\small
\centering
\scalebox{0.8}{%

\begin{tabular}{l|l|l|ccc|ccc}
                                                 & method & backbone & $AP$ & $AP_{50}$ & $AP_{75}$ & $AP_S$ & $AP_M$ & $AP_L$ \\ \hline
\multicolumn{1}{c|}{\multirow{4}{*}{Two-stage}} &Faster R-CNN w/ TDM \cite{tdm} & Inception-ResNet-v2-TDM & 36.8 & 57.7 & 39.2 & 16.2 & 39.8 & 52.1 \\
\multicolumn{1}{c|}{}                            &Faster R-CNN w/ FPN \cite{fpn} & ResNet-101-FPN & 36.2 & 59.1 & 39.0 & 18.2 & 39.0 & 48.2 \\
\multicolumn{1}{c|}{}                            &Faster R-CNN by G-RMI \cite{speedAccuracy} & Inception-ResNet-v2 & 34.7 & 55.5 & 36.7 & 13.5 & 38.1 & 52.0 \\
\multicolumn{1}{c|}{}                            &Faster R-CNN w/ TDM \cite{tdm} & Inception-ResNet-v2-TDM & 36.8 & 57.7 & 39.2 & 16.2 & 39.8 & 52.1 \\ \hline \hline
\multicolumn{1}{l|}{\multirow{6}{*}{\begin{tabular}[c]{@{}c@{}}Slow \\ Single-stage\end{tabular}}} 
						&SSD-513 \cite{ssd} & ResNet-101-SSD & 31.2 & 50.4 & 33.3 & 10.2 & 34.5 & 49.8 \\
\multicolumn{1}{l|}{}   &DSSD-513 \cite{dssd} & ResNet-101-DSSD & 33.2 & 53.3 & 35.2 & 13.0 & 35.4 & 51.1\\
\multicolumn{1}{l|}{} 	&RetinaNet-800 \cite{retina} & ResNet-101-FPN & 39.1 & 59.1 & 42.3 & 21.8 & 42.7 & 50.2\\
\multicolumn{1}{l|}{}   &RetinaNet-800 \cite{retina} & ResNeXt-32x8d-101-FPN & 40.8 & 61.1 & 44.1 & 24.1 & 44.2 & 51.2\\ 
\multicolumn{1}{l|}{}   &CornerNet \cite{cornernet} & Hourglass-104 & 40.5 & 56.5 & 43.1 & 19.4 & 42.7 & \textbf{53.9} \\
\multicolumn{1}{l|}{}   &FCOS-800 \cite{fcos} & ResNet-101-FPN & 41.0 & 60.7 & 44.1 & 24.0 & 44.1 & 51.0 \\
\multicolumn{1}{l|}{}   &FCOS-800 \cite{fcos} & ResNeXt-32x8d-101-FPN & \textbf{42.1} & \textbf{62.1} & \textbf{45.2} & \textbf{25.6} & \textbf{44.9} & 52.0 \\ 
\hline \hline
\multicolumn{1}{l|}{\multirow{7}{*}{\begin{tabular}[c]{@{}c@{}}Fast \\ Single-stage\end{tabular}}} 
												 &SSD-321 \cite{ssd} & ResNet-101-SSD & 28.0 & 45.4 & 29.3 & 6.2 & 28.3 & 49.3 \\
\multicolumn{1}{l|}{}							 &DSSD-321 \cite{dssd} & ResNet-101-DSSD & 28.0 & 46.1 & 29.2 & 7.4 & 28.1 & 47.6 \\
\multicolumn{1}{l|}{}                            &RetinaNet-500 \cite{retina} & ResNet-50-FPN & 32.5 & 50.9 & 34.8 & 13.9 & 35.8 & 46.7 \\ 
\multicolumn{1}{l|}{}                            &RetinaNet-500 \cite{retina} & ResNet-101-FPN & 34.4 & 53.1 & 36.8 & 14.7 & 38.5 & 49.1 \\
\multicolumn{1}{l|}{}                            &YoloV2-544 \cite{yolov2} & DarkNet-19 & 21.6 & 44.0 & 19.2 & 5.0 & 22.4 & 35.5 \\ 
\multicolumn{1}{l|}{}                            &YoloV3-608 \cite{yolov3} & DarkNet-53 & 33.0 & \textbf{57.9} & 34.4 & 18.3 & 35.4 & 41.9 \\ \cline{2-9}
\multicolumn{1}{l|}{} 							 & Lapnet-512 & DarkNet-53 & 37.6 & 55.5 & 40.4 & 17.6 & 40.5 & \textbf{49.9} \\ 
\multicolumn{1}{l|}{} 							 & Lapnet-608 & DarkNet-53 & \textbf{38.2} & 56.6 & \textbf{41.2} & \textbf{20.3} & \textbf{41.6} & 47.5 
\end{tabular}}
\vspace*{2mm}
\caption{LapNet results on MSCOCO \texttt{test-dev2017} with two different input resolutions (512x512 and 608x608). They are compared to results from several previous methods. LapNet clearly outperforms previous fast single-stage detectors. Figure \ref{fig:tradeoff} presents these results in the form of speed/accuracy graph.}
\label{results_coco}
\end{table*}

\section{Conclusion}

We have introduced LapNet, a real time single-shot detector which gives a very good trade-off between speed and accuracy. The training process is based on a new positive/negative label assignment method using Per-Object Normalized Overlap and Ambiguity Management Strategy. An automatic weighting method is also proposed to balance losses, classes and object sizes for efficient learning without any extra cost in computation time. Experiments on two reference datasets demonstrate the relevance of our contributions and position LapNet as the most accurate detector among real-time methods. We believe that the proposed generic automatic balancing could be applied to many other computer vision tasks.

\bibliographystyle{splncs04}
\bibliography{egbib}
\end{document}